\documentclass[10pt,twocolumn,letterpaper]{article}

\usepackage[pagenumbers]{cvpr} % To force page numbers, e.g. for an arXiv version

\usepackage{multirow} 
\definecolor{cvprblue}{rgb}{0.21,0.49,0.74}
\usepackage[pagebackref,breaklinks,colorlinks,allcolors=cvprblue]{hyperref}

\def\confName{CVPR}
\def\confYear{2026}

\title{VSDiffusion: Taming Ill-Posed Shadow Generation via Visibility-Constrained Diffusion}

\author{
Jing Li \qquad Jing Zhang$^*$\\
East China University of Science and Technology, Shanghai, China\\
{\tt\small lij1@mail.ecust.edu.cn \qquad jingzhang@ecust.edu.cn}
}
\begin{document}
\maketitle
\begin{abstract}
 Generating realistic cast shadows for inserted foreground objects is a crucial yet challenging problem in image composition, where maintaining geometric consistency of shadow and object in complex scenes remains difficult due to the ill-posed nature of shadow formation. To address this issue, we propose VSDiffusion, a visibility-constrained two-stage framework designed to narrow the solution space by incorporating visibility priors. In Stage I, we predict a coarse shadow mask to localize plausible shadow generated regions. And in Stage II, conditional diffusion is performed guided by lighting and depth cues estimated from the composite to generate accurate shadows. In VSDiffusion, we inject visibility priors through two complementary pathways. First, a visibility control branch with shadow-gated cross attention that provides multi-scale structural guidance. Then, a learned soft prior map that reweights training loss in error-prone regions to enhance geometric correction. Additionally, we also introduce high-frequency guided enhancement module to sharpen boundaries and improve texture interaction with the background. Experiments on widely used public DESOBAv2 dataset demonstrated that our proposed VSDiffusion can generate accurate shadow, and establishes new SOTA results across most evaluation metrics.
\end{abstract}    
\section{Introduction}
\label{sec:intro}

“The beginnings and ends of shadow lie between light and darkness.”--da Vinci.

In recent years, diffusion models\cite{ho2020denoising, rombach2022high} and their conditional control techniques \cite{zhang2023adding} have greatly advanced image synthesis, enabling high-resolution, realistic, and controllable generation. However, in real-world applications such as film production and e-commerce design, where a foreground object must be composited into a specific background, the realism of the generated image often depends on physical consistency. If the foreground shadow is missing, has an incorrect direction, or exhibits an implausible shape, the composite can look unnatural even when the object itself is well aligned. As noted by Hu \etal~\cite{hu2024unveiling}, shadows are an indispensable component of image compositing and are critical for scene understanding and editing consistency. Therefore, generating shadows that are geometrically correct and perceptually realistic is essential for improving composited image quality, and it is the goal of this work.

Existing shadow generation methods can mainly be divided into two categories: rendering-based methods\cite{sheng2021ssn,sheng2022controllable,sheng2023pixht} and non-rendering methods \cite{hong2022shadow,liu2024shadow,zhao2025shadow}. The former rely on strong assumptions such as accurate lighting and material properties, which are hard to satisfy in real workflows. The latter learn the mapping between the input and the shadow in a data-driven manner and have become the mainstream direction. Early work such as SGRNet\cite{hong2022shadow} constructed the DESOBA dataset and used a two-stage pipeline with geometric localization followed by appearance completion. SGDiffusion\cite{liu2024shadow} introduced this task into the diffusion framework and expanded the dataset to DESOBAv2 to improve shadow quality. GPSDiffusion\cite{zhao2025shadow} further added geometric priors, such as shape prototypes, to refine shadow shape and placement. In the above training and inference settings of shadow generation model, the model only observes a composited image and limited shadow supervision (\eg, a binary mask), while key physical information, such as the precise light distribution and scene geometry, is missing. This results in the same input corresponding to multiple visually plausible shadow outputs, and learning can degenerate into fitting local image textures, which makes it hard to guarantee geometrically correct shadows.

%Although recent methods have significantly improved visual fidelity, cast shadow generation in complex scenes remains challenging. Common failures—such as incorrect shadow directions, misaligned contact regions, and occlusion-inconsistent shadows—suggest that the problem stems from the intrinsic structural ambiguity of shadow formation rather than insufficient model capacity.
                                                                                                           
\begin{figure*}[t]
  \centering
  \includegraphics[width=0.85\textwidth]{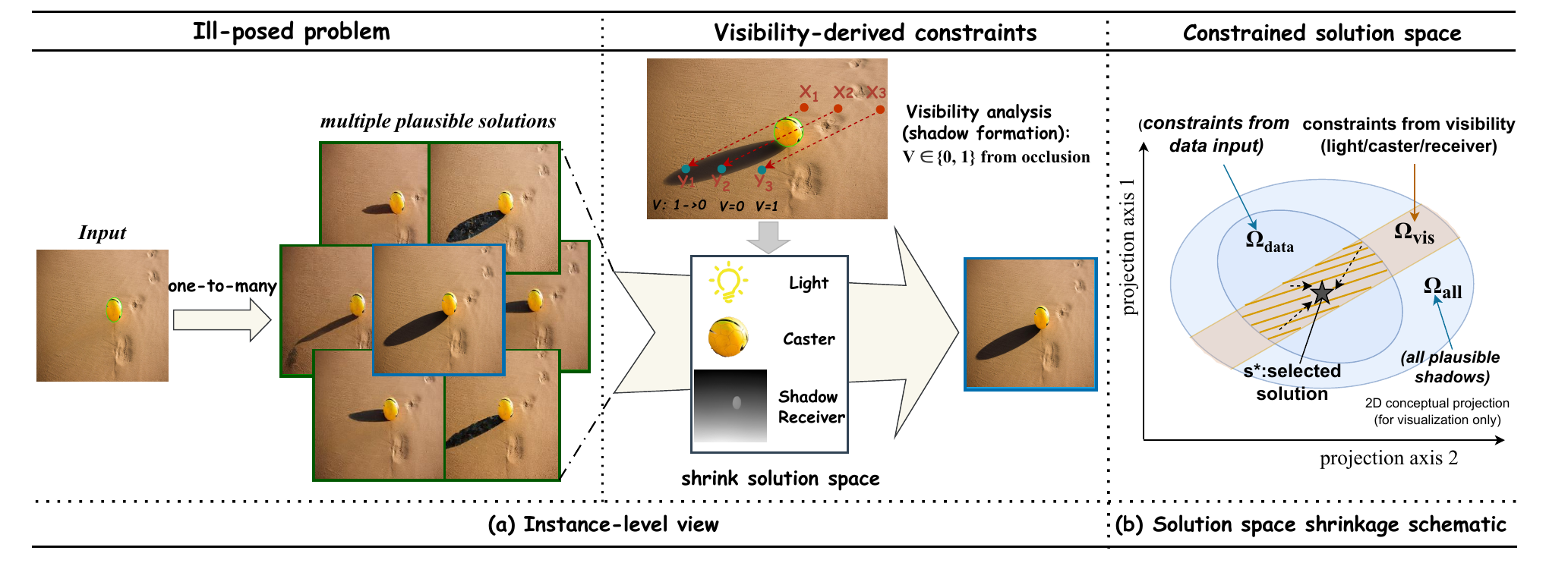}
  \caption{(a) Shadow generation is an ill-posed problem: one input may correspond to multiple visually plausible shadows. With visibility analysis, geometric constraints from the light source, caster, and the shadow receiver can significantly narrow the solution space. (b) further illustrates how the solution space is progressively narrowed. The hatched area indicates the intersection of $\Omega_{vis}$, $\Omega_{data}$, and $\Omega_{all}$. S* is the selected solution that balances data fidelity with geometric plausibility.
  }
  \label{fig:f1}
\end{figure*}

From a computational and mathematical perspective, shadow generation is inherently ill-posed \cite{kabanikhin2011inverse}, as multiple plausible shadows can correspond to the same scene, and small changes in scene configuration can lead to large variations in shadow appearance.  From a physical standpoint, a shadow is an interpretable phenomenon caused by occlusion: when light rays are blocked, visibility changes \cite{dutre2003global,cohen1993radiosity}. The idea of visibility editing in graphics \cite{obert2010visibility} further shows that explicitly controlling visibility, decoupled from illumination and materials, allows shadow geometry to be manipulated precisely.

Motivated by this insight, we try to revisit shadow generation from a visibility-centric viewpoint and settle it by narrowing the solution space of ill-posed problems. Without relying on full physical simulation, we use available physical cues to approximately infer visibility changes in the scene. These cues act as strong constraints and substantially shrink the solution space. Under this perspective, we prioritize geometric plausibility, (especially layout consistency and lighting direction), and then refine appearance details, providing a more robust starting point for method design of shadow generation.

Based on above analysis, we formalize the problem as illustrated in Fig. 1 \cref{fig:f1}. Because the input (a composite image) lacks precise physical quantities, shadow generation is a typical one-to-many mapping with a large solution space, which is a classical ill-posed problem. We then analyze shadow formation through visibility: a shadow arises when the visibility between a light point $x$ and a receiver point $y$ is blocked by intermediate geometry\cite{laine2006incremental}. If the path is unobstructed, $V(x,y)=1$ and y is not in shadow; if it is fully blocked, $V(x,y)=0$ and y lies in shadow. The transition of V from 1 to 0 defines the shadow boundary. Therefore, shadow geometry is mainly determined by the spatial relationship among the light source, the caster, and the shadow receiver \cite{haines2019ray,woo2002survey}, and we introduce these factors as constraints in the generation process to narrow the solution space.

Based on the above analysis, we propose visibility-constrained shadow diffusion model(VSDiffusion), in which visibility-derived priors are injected into the diffusion process in two complementary ways. First, we extract prior features with a Visibility Control Branch (VCB) and design a Shadow-Gated Cross Attention (SGCA) module, which injects conditional features into denoising in a multi-scale and sparsely localized manner to provide controllable structural guidance. Second, we adopt a lightweight U-Net \cite{ronneberger2015u} to predict a visibility-guided spatial weight map, which reweights the training loss to emphasize error-prone shadow regions and improve geometric alignment. Moreover, to address blurry shadow contours and weak texture interactions, we introduce High-Frequency Guided Enhancement (HFGE) module to refine boundary transitions and improve texture fusion with the background, further enhancing perceptual realism. The main contributions of this paper are listed as follows:

\begin{itemize}

   \item We formalized the shadow generation as a ill-posed problem and proposed a two-stage shadow generation model that advances shadow generation from purely data-driven approach toward visibility priors guided framework, thus effectively reducing the solution space and improving geometric consistency of generated shadows.   

   \item We integrated two complementary forms of prior injection: structured guidance via SGCA module during denoising, and spatially weighted optimization constraints that focus learning on geometrically critical regions. This combination effectively enhances geometric plausibility.   

   \item We introduced HFGE module to strengthen high-frequency detail modeling, improving shadow edge quality and the realism of background texture interactions.

\end{itemize}
\section{Related Work}
\label{sec:relat}

In this section, we review some methods that are closely related to our research, including image composition and shadow generation.
\subsection{Image Composition}

Image composition \cite{niu2106making} aims to place a foreground object into a background and generate a realistic composite image. Early methods treated it as a signal processing problem. They mainly reduced low-level discontinuities and boundary artifacts. The Porter--Duff operator \cite{porter1984compositing} provides an alpha-based linear compositing model that standardizes image overlay. Poisson image editing \cite{perez2003poisson} further enforces seamless boundaries via gradient-domain optimization. However, these methods are mostly local smoothing. They do not model high-level semantics or illumination. In the deep learning era, image composition shifted from pixel-domain optimization to data-driven semantic consistency modeling. Many methods use encoder--decoder backbones, such as U-Net \cite{ronneberger2015u}. They can also adopt conditional GANs \cite{isola2017image} with adversarial training to improve structural and textural realism. This line of work has branched into more focused directions, including learning-based image blending \cite{wu2019gp,zhang2019all}, object placement assessment \cite{liu2021opa,niu2022fast}, and image harmonization \cite{chen2023hierarchical, chen2023dense,cong2020dovenet}. Despite better global statistics and color matching, these methods often produce semantic mismatches and structural artifacts near occlusion boundaries. A key reason is the difficulty of modeling complex 3D interactions using only 2D feature maps. More recently, diffusion-model-based generative methods \cite{chen2024anydoor,kulal2023putting,sarukkai2024collage} shifted composition from patching pixels to regenerating content under strong priors. Composition is no longer simple layer overlaying. Instead, it becomes an iterative denoising process conditioned on the background and an initial mask. These methods improve visual realism. However, their priors favor plausible appearances. It is still hard to guarantee both physical consistency and fine-grained structural fidelity in complex scenes.

\subsection{Shadow Generation}

Shadow generation is crucial for the realism and spatial coherence of composite images. Existing methods mainly follow two paradigms. Rendering-based methods synthesize shadows using explicit geometry or interpretable representations, requiring accurate geometric input. Classical techniques like shadow mapping \cite{williams1978casting,liu2009shadow} and its variants \cite{stamminger2002perspective,dou2014adaptive,donnelly2006variance} enable real-time rendering but suffer from artifacts. Recent advances integrate neural rendering for improved control and realism. For instance, SSN \cite{sheng2021ssn} generates interactive soft shadows from 2D masks, using ambient light to enhance consistency. Methods like Pixel Height Maps \cite{sheng2022controllable} and PixHt-Lab \cite{sheng2023pixht} incorporate per-pixel height to create a projectable 3D representation, moving beyond pure 2D synthesis.

Non-rendering methods learn shadow mappings directly from paired shadow and shadow-free images. Early GAN-based methods include ShadowGAN \cite{zhang2019shadowgan} and ARShadowGAN \cite{liu2020arshadowgan}. ShadowGAN \cite{zhang2019shadowgan} introduced conditional GANs for shadow generation and improved visual quality. ARShadowGAN \cite{liu2020arshadowgan} uses attention and known background shadows as references to model the relation between foreground shadows and scene context. In addition, some shadow removal methods can also provide useful cues for shadow generation, such as Mask-ShadowGAN \cite{hu2019mask}. Two-stage networks such as SGRNet \cite{hong2022shadow} decouple shadow region detection and shading for more reliable results. DMASNet \cite{tao2024shadow} further decomposes mask prediction into box-level localization and shape refinement, improving boundary quality and generalization in complex scenes. With the development of diffusion models, SGDiffusion \cite{liu2024shadow} improves controllability of shadow appearance via conditional control branches and intensity modulation. GPSDiffusion \cite{zhao2025shadow} introduces geometric priors (\eg, shape prototypes) to constrain shadow position and shape. Despite these advances, data-driven methods with implicit physical constraints still often fail to ensure plausible shadows in complex scenes.

\paragraph{In summary.}
General image composition methods mainly optimize appearance consistency and often leave cast-shadow geometry implicit. Dedicated shadow generation methods face a trade-off: rendering-based approaches lack generalization, while data-driven approaches lack physical constraints. These gaps motivate incorporating visibility-derived constraints into diffusion-based generation to shrink the feasible solution space and improve geometric consistency without full physical simulation. 
\section{Method}

We first present an overview of the proposed VSDiffusion framework, and then describe each major module in detail.

\subsection{Overview}

\begin{figure*}[t]
  \centering
  \includegraphics[width=0.98\textwidth]{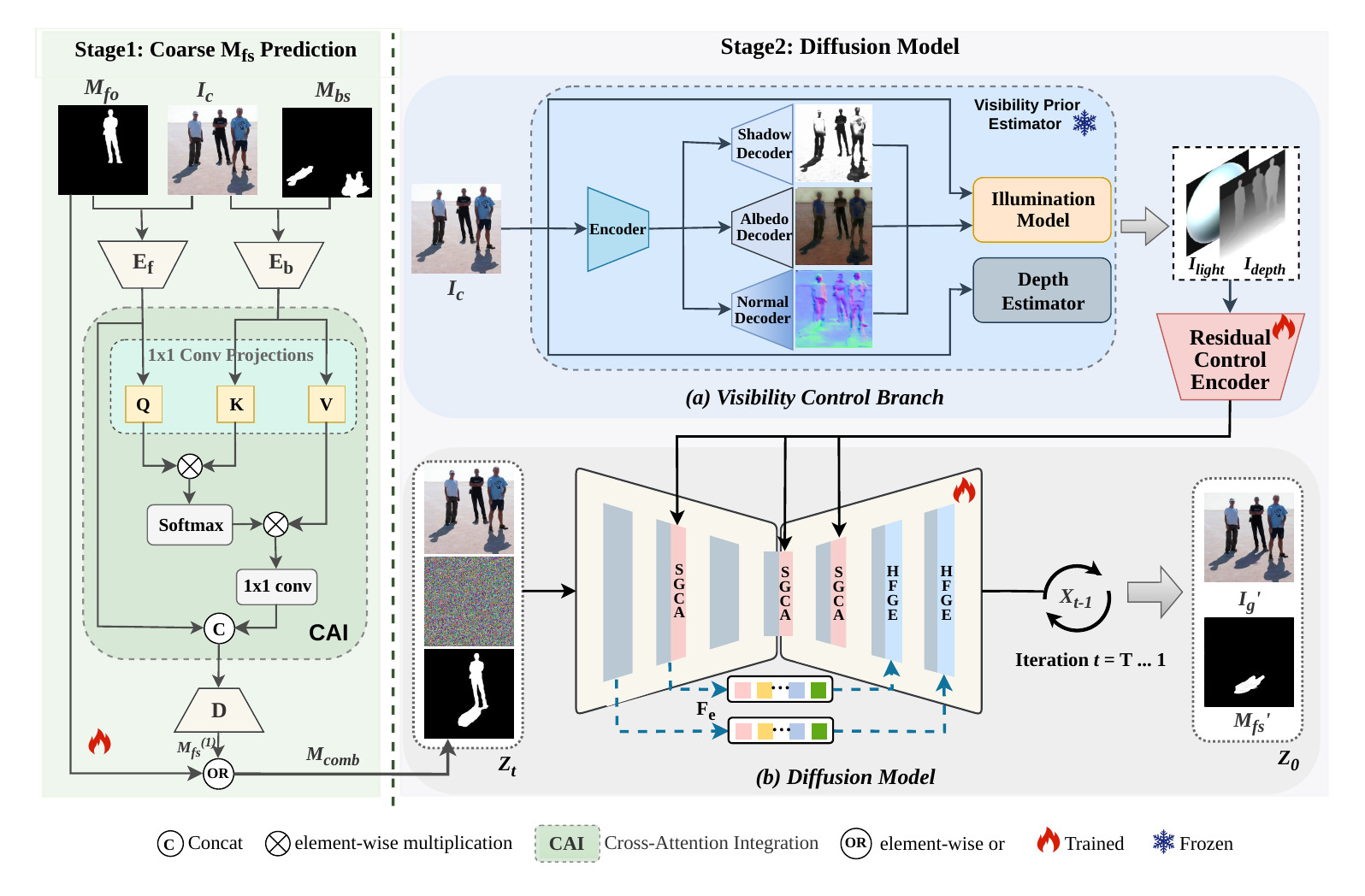}
  \caption{Framework of our VSDiffusion. It consists of two stages. Stage 1 predicts a coarse foreground shadow mask $M_{fs}^{(1)}$. Stage 2 includes two sub-modules: (a) the Visibility Control Branch (VCB), which utilizes visibility prior estimator to extract $I_{\text{light}}$ and $I_{\text{depth}}$ from $I_\text{c}$, subsequently encoded by RCE to provide structural guidance. (b) Under this guidance, the U-Net progressively denoises $Z_t$ into $Z_0$. 
  }
  \label{fig:f2}
\end{figure*}

Our method aims to generate physically consistent shadows for foreground objects in composited images. Given a composite image $I_\text{c}$ where the inserted foreground object lacks its corresponding shadow, along with a foreground mask $M_{fo}$ and a background shadow mask $M_{bs}$, the goal is to infer plausible shadow. This problem is ill-posed \cite{kabanikhin2011inverse}, leading to non-unique and ambiguous solutions. To mitigate this ambiguity, we propose a two-stage framework that progressively constrains the solution space as shown in \cref{fig:f2}. 

In the first stage, we predict a coarse foreground shadow mask $M_{fs}^{(1)}$, which serves as a spatial prior and reduces geometric uncertainty. Specifically, the background encoder $E_b$ and foreground encoder $E_f$ process $(I_c, M_{bs})$ and $(I_c, M_{fo})$, respectively. Their features interact via cross-attention integration \cite{hong2022shadow} to produce $M_{fs}^{(1)}$, which is then combined with $M_{fo}$ to form $M_{\text{comb}}$. In the second stage, we incorporate visibility priors into the diffusion process to further regularize the shadow generation and alleviate the inherent ill-posedness. These priors guide the denoising U-Net. At each timestep, the U-Net denoises $(I_c, M_{\text{comb}}, x_t)$ and outputs the final shadowed image $I_g^{\prime}$.

In this work, these visibility priors are injected through two complementary pathways: (i) As structurally controllable conditional guidance, VCB injects priors via SGCA, which sparsely injects guidance at three U-Net scales to align geometry, preserve global layout, and refine boundaries. (ii) As a training-level spatial supervision signal, we introduce $S_{\text{prior}}$-Weighted Loss (SWL), which uses a soft prior map to spatially reweight pixel-level supervision to focus on error-prone regions. These two components are complementary, providing both architectural constraints and adaptive optimization guidance. In addition, HFGE extracts high-frequency cues from shallow encoder layers and residually injects them during high-resolution decoding to enhance shadow boundaries and texture details. 
The VCB is introduced in \cref{sec:VCB}. SGCA and HFGE are detailed in \cref{sec:denoiser}. The SWL term is described in \cref{sec:trainloss}.

\subsection{Visibility Control Branch}
\label{sec:VCB}
Predicting shadows from a single input image $I_\text{c}$ remains highly ambiguous without explicit constraints on scene structure and illumination, often yielding multiple visually plausible yet non-unique solutions. Therefore, we propose the Visibility Control Branch (VCB) to guide the denoising network by incorporating visibility-derived constraints, thereby narrowing the solution space. As shown in \cref{fig:f2} (a), the VCB comprises two components: visibility prior estimator \cite{yu2021outdoor,ranftl2020towards} and Residual Control Encoder \cite{kocsis2024lightit}.

The visibility prior estimator is designed to extract visibility-related priors (light and depth) from $I_\text{c}$. For illumination estimation, it employs a shared encoder followed by three lightweight decoders to predict shadow, albedo, and normal maps. These estimated attributes, together with $I_\text{c}$, are then fed into an illumination model based on Lambertian reflectance \cite{forsyth2002computer, basri2003lambertian}. This model recovers the second-order spherical harmonics coefficients representing global illumination, which are visualized as an illumination map $I_{\text{light}}$ to depict light direction and intensity. Since the illumination $I$ can be derived analytically from the input image and the estimated albedo $\alpha$, shadow weight $s$, and surface normal $n$, no separate network branch is required for lighting prediction. For a pixel $(x,y)$, the image formation process is formulated as:
\begin{equation}
  i(x,y) = \alpha(x,y) \odot s(x,y)\, B\bigl(n(x,y)\bigr)\, I
  \label{eq:illum_model}
\end{equation}
where $i(x,y)$ is the pixel value in $I_\text{c}$, $\odot$ denotes element-wise multiplication, $B(n)$ denotes the second-order SH basis functions evaluated at the surface normal $n$, and $I$ denotes the unknown second-order SH coefficients, \ie $I_{\text{light}}$. By formulating this as a least-squares problem, $I$ can be solved via inversion. Overall, $I_{\text{light}}$ is obtained following an inverse-rendering framework \cite{yu2021outdoor}. Additionally, the visibility prior estimator utilizes a pre-trained MiDaS model \cite{ranftl2020towards} as a monocular depth estimator to produce a depth map $I_{\text{depth}}$ from $I_\text{c}$.
 
To stably integrate these visibility priors, we employ a Residual Control Encoder \cite{kocsis2024lightit} for feature extraction. This encoder adopts a lightweight residual design (7 blocks) with zero convolutions, in contrast to the encoder in ControlNet \cite{zhang2023adding}, which may ignore control signals during early training and cause instability. This design ensures that the diffusion process gradually incorporates $I_{\text{light}}$ and $I_{\text{depth}}$, improving geometric and illumination alignment.

\subsection{Geometry-Detail Denoiser}
\label{sec:denoiser}
\subsubsection{Shadow-Gated Cross Attention Module}
\label{sec:sgca}
To effectively exploit visibility priors, we propose Shadow-Gated Cross-Attention (SGCA) module. Unlike conventional dense conditioning across many layers \cite{zhang2023adding}, SGCA injects these priors ($I_{\text{light}}, I_{\text{depth}}$) into three strategically selected U-Net anchors (early, mid, and late) for geometric alignment, global layout, and refinement. Dense injection maybe restrict the generation space, compromise consistency, and cause over-conditioning. By injecting guidance at these anchors, SGCA maintains controllability while avoiding over-conditioning and redundant computation, offering a better trade-off between conditioning strength and generative fidelity. Here, sparse refers to the sparsity of injection locations rather than token-level attention sparsity.
\begin{figure}[!t]
  \centering
  \includegraphics[width=\linewidth]{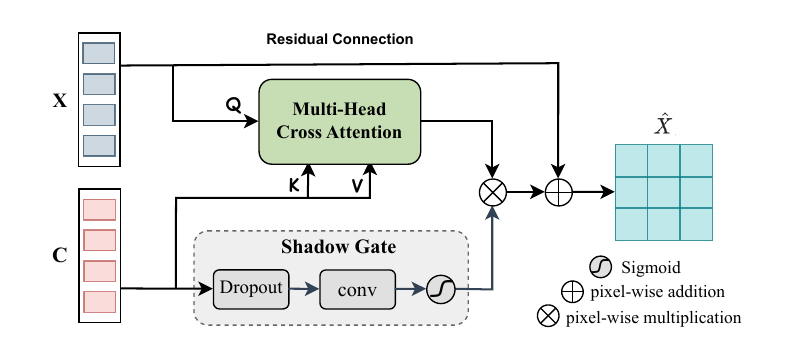}
  \caption{Architecture of the Shadow-Gated Cross-Attention (SGCA) module. SGCA computes cross-attention between U-Net features $X$ (queries) and conditional features $C$ (keys/values). Shadow Gate mechanism, consisting of dropout, convolution, and sigmoid, adaptively modulates the attention output before it is residually added to $X$, ensuring robust and selective integration of visibility priors.
  }
  \label{fig:f3}
\end{figure}
As shown in \cref{fig:f3}, let $X^{(s)}\in\mathbb{R}^{B\times C_s\times H_s\times W_s}$ denote the U-Net feature at scale $s$, and $C^{(s)}$ represent the aligned conditional feature from the Residual Control Encoder. SGCA computes cross-attention where the queries $Q$ are projected from $X^{(s)}$, and keys $K$ and values $V$ are derived from $C^{(s)}$. To adaptively regulate the influence of external priors, a Shadow Gate $G^{(s)}$ is predicted from $C^{(s)}$ by a lightweight head $\psi(\cdot)$:
\begin{equation}
G^{(s)}=\sigma\!\left(\psi\!\left(C^{(s)}\right)\right)
\end{equation}
\begin{equation}
\hat{X}^{(s)} = X^{(s)} + G^{(s)} \odot \mathrm{unflat}\!\left(A^{(s)}\right)
\end{equation}
where $\sigma$ is the sigmoid function, $\operatorname{unflat}(\cdot)$ restores the attention tokens to spatial dimensions, and $A^{(s)} = MHA(Q, K, V)$. The gate strengthens the injection when conditional features are beneficial for shadow inference and suppresses them otherwise, preventing texture degradation and artifact amplification.

\subsubsection{High-Frequency Guided Enhancement}
\label{sec:hfge}
Shadow generation often suffers from blurry boundaries, jagged artifacts, and over-smoothed background textures. To address these issues, we introduce High-Frequency Guided Enhancement (HFGE). HFGE extracts robust high-frequency cues from shallow U-Net encoder features and injects them as residual guidance into the late decoder stages, ensuring sharp shadow edges without compromising background fidelity.

We extract high-frequency information from shallow encoder features $F_e$ based on two key insights. First, shallow features contain mixed-frequency signals, including local gradients and texture responses, which are suitable for describing edges and fine details. Second, in diffusion models, high-frequency details are typically refined in later steps. As observed by Falck \etal~\cite{falck2025fourier}, high-frequency components enter the low Signal-to-Noise Ratio (SNR) regime earlier and faster during the DDPM forward process. Low-frequency structures tend to appear first, while high-frequency details emerge gradually at later stages. Thus, injecting guidance at late high-resolution decoding stages aligns with this natural refinement order. Conversely, injecting too early, \eg, in the encoder or bottleneck, risks feature dilution through downsampling or introducing noise that degrades global shadow formation.
\begin{figure*}[t]
  \centering
  \includegraphics[width=0.85\textwidth]{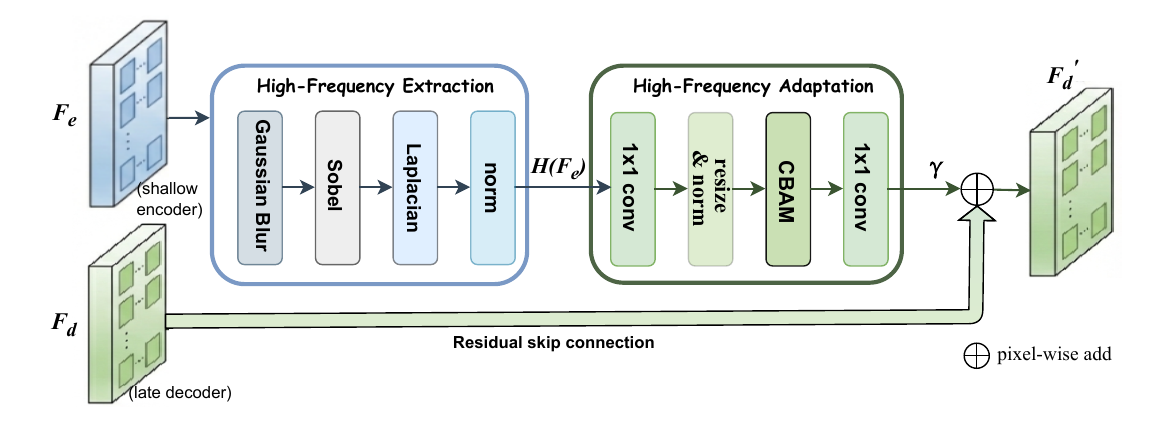}
  \caption{Architecture of the High-Frequency Guided Enhancement (HFGE) module. HFGE consists of High-Frequency Extraction (blue) to capture structural cues from $F_e$, and High-Frequency Adaptation (green) to align and calibrate these signals via CBAM. The refined guidance is residually added to decoder features $F_d$ to enhance shadow boundaries and textures.
  }
  \label{fig:f4}
\end{figure*}

As shown in \cref{fig:f4}, the HFGE module consists of two main stages: High-Frequency Extraction (blue region) and High-Frequency Adaptation (green region). For High-Frequency Extraction, we extract robust high-frequency cues from shallow encoder features $F_e$ through four sequential steps. First, we apply spatial Gaussian smoothing to $F_e$ to filter out local pseudo-textures, producing $\tilde{F}_e$. Second, we use Sobel filters to compute gradients $g_x$ and $g_y$, obtaining the gradient magnitude $G$ to emphasize structural boundaries:
\begin{equation}
G=\sqrt{g_x^{2}+g_y^{2}+\epsilon}
\end{equation}
Third, to further capture fine-grained structures like thin edges, we incorporate the response $L$ from the Laplacian operator. Finally, we normalize the magnitude over spatial dimensions to obtain the high-frequency feature $H(F_e)$:
\begin{equation}
H(F_e)=\mathrm{Norm}(G+\alpha L)
\label{eq:hfge_highfreq}
\end{equation}
where $\alpha$ is a weighting factor and $\mathrm{Norm}(\cdot)$ denotes spatial magnitude normalization. For High-Frequency Adaptation, since $H(F_e)$ and decoder features $F_d$ reside in different feature spaces, we introduce an adaptation branch to align the signals. Specifically, we employ a $1\times1$ convolution and bilinear interpolation to match the channel dimensions and spatial resolution of $F_d$. Then, a lightweight CBAM \cite{woo2018cbam} is applied to adaptively reweight the responses, focusing on critical shadow contours. The calibrated guidance is then residually added to $F_d$:
\begin{equation}
F_d' = F_d + \gamma \cdot \mathrm{CBAM}\big(\phi(H(F_e))\big)
\label{eq:hfge_inject}
\end{equation}
where $\phi(\cdot)$ denotes projection and scale alignment, and $\gamma$ controls the injection strength. This design allows high-frequency details to refine boundaries while preserving the underlying background texture.

\subsection{Training Losses}
\label{sec:trainloss}
We adopt a three-step training strategy: (i) pretraining Stage I alone to learn basic geometric localization; (ii) training Stage II for shadow synthesis with Stage I frozen; and (iii) joint fine-tune the full framework at a lower learning rate. We optimize the model with the following loss functions.
\subsubsection{First Stage Loss}
The coarse foreground shadow mask $M_{\text{fs}}^{(1)}$ is supervised by a combination of Binary Cross-Entropy (BCE) and Dice loss against the ground truth $M_{\text{fs}}$:
\begin{equation}
\mathcal{L}_{s1} = \mathcal{L}_{\mathrm{bce}}\big(M_{fs}^{(1)}, M_{fs}\big)
+ \mathcal{L}_{\mathrm{dice}}\big(M_{fs}^{(1)}, M_{fs}\big)
\label{eq:s1}
\end{equation}

\begin{table*}[!t]
  \caption{Quantitative comparisons on DESOBAv2. We evaluate shadow generation with RMSE, SSIM, and BER. BOS denotes test images with background shadow references. $\uparrow$/$\downarrow$ indicate higher/lower is better. Best results are highlighted in bold.
  }
  \label{tab:quantitative_results}
  \centering
  \setlength{\tabcolsep}{3pt} % 减小列之间的默认间距，大幅压缩总宽度
  \resizebox{0.8\textwidth}{!}{ % 自动缩放到页面宽度，彻底解决溢出
  \begin{tabular}{@{}l|cccccc|cccccc@{}}
    \toprule
    \multirow{2}{*}{Method} & \multicolumn{6}{c|}{BOS Test Images} & \multicolumn{6}{c}{BOS-free Test Images} \\
    & GR$\downarrow$ & LR$\downarrow$ & GS$\uparrow$ & LS$\uparrow$ & GB$\downarrow$ & LB$\downarrow$ & GR$\downarrow$ & LR$\downarrow$ & GS$\uparrow$ & LS$\uparrow$ & GB$\downarrow$ & LB$\downarrow$ \\
    \midrule
    ShadowGAN\cite{zhang2019shadowgan} & 7.511 & 67.464 & 0.961 & 0.197 & 0.446 & 0.890 & 17.325 & 76.508 & 0.901 & 0.060 & 0.425 & 0.842 \\
    Mask-SG\cite{hu2019mask} & 8.997 & 79.418 & 0.951 & 0.180 & 0.500 & 1.000 & 19.338 & 94.327 & 0.906 & 0.044 & 0.500 & 1.000 \\
    AR-SG\cite{liu2020arshadowgan} & 7.335 & 58.037 & 0.961 & 0.241 & 0.383 & 0.682 & 16.067 & 63.713 & 0.908 & 0.104 & 0.349 & 0.682 \\
    SGRNet\cite{hong2022shadow} & 7.184 & 68.255 & 0.964 & 0.206 & 0.301 & 0.596 & 15.596 & 60.350 & 0.909 & 0.100 & 0.271 & 0.534 \\
    DMASNet\cite{tao2024shadow} & 8.256 & 59.380 & 0.961 & 0.228 & 0.276 & 0.547 & 18.725 & 86.694 & 0.913 & 0.055 & 0.297 & 0.574 \\
    SGDiffusion\cite{liu2024shadow} & 6.098 & 53.611 & \textbf{0.971} & 0.370 & 0.245 & 0.487 & 15.110 & 55.874 & 0.913 & 0.117 & 0.233 & 0.452 \\
    GPSDiffusion\cite{zhao2025shadow} & \textbf{5.896} & 46.713 & 0.966 & 0.374 & 0.213 & 0.423 & \textbf{13.809} & 55.616 & \textbf{0.917} & 0.166 & 0.197 & 0.384 \\
    Ours & 6.422 & \textbf{46.542} & 0.964 & \textbf{0.377} & \textbf{0.182} & \textbf{0.360} & 14.629 & \textbf{55.377} & 0.912 & \textbf{0.184} & \textbf{0.194} & \textbf{0.373} \\
    \bottomrule
  \end{tabular}
  }
\end{table*}

\subsubsection{Second Stage Loss}
Stage II focuses on high-fidelity shadow reconstruction and mask refinement via two terms. 

\noindent 1) Base Loss: To ensure pixel-level consistency for the generated image $\tilde{I}$ and refined mask $\tilde{M}_{fs}$, we define the base loss as:
\begin{equation}
\begin{aligned}
\mathcal{L}_{\text{base}}
&= \frac{1}{N}\sum_{x,y} \Big(
\|\tilde{I}(x,y)-I(x,y)\|_1 \\
&\qquad\qquad + \lambda_1 \|\tilde{M}_{fs}(x,y)-M_{fs}(x,y)\|_2^2
\Big)
\end{aligned}
\label{eq:base}
\end{equation}
where $I$ is the ground truth shadowed image, and $N$ is the pixel count.

\noindent 2) $S_{\text{prior}}$-Weighted Loss (SWL): Shadow errors tend to focus on small but critical regions, such as thin shadow edges or misaligned boundaries. Standard global loss often dilute gradient in these critical areas. Thus, we introduce a soft prior map $S_{\text{prior}}$ to spatially re-weight the optimization process. 

As described in \cref{fig:f5}, this map is generated by a lightweight U-Net network $G_{\text{p}}$, which aggregates visibility-related cues and maps the output to [0,1] via a sigmoid function:

\begin{figure}[!ht]
  \centering
  \includegraphics[width=0.98\linewidth]{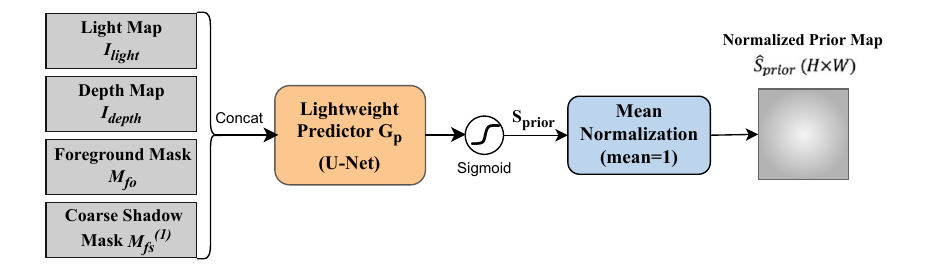}
  \caption{Generation of the spatial prior in SWL. Visibility-related inputs $\{I_{\text{light}}, I_{\text{depth}}, M_{\text{fo}}, M_{\text{fs}}^{(1)}\}$ are concatenated and fed into a lightweight predictor $G_p$, followed by sigmoid to obtain $S_{\text{prior}}$. Mean normalization then produces $\hat{S}_{\text{prior}}$ (with mean 1), which is used for spatial reweighting in SWL.
  }
  \label{fig:f5}
\end{figure}

\begin{equation}
S_{\text{prior}} = \sigma\!\left( G_{p}\!\left(
I_{\mathrm{light}}, I_{\mathrm{depth}}, M_{fo}, M_{fs}^{(1)}\right)\right),
\quad
S_{\text{prior}} \in [0, 1]
\end{equation}
where $G_{\text{p}}$ is the lightweight $S_{\text{prior}}$ predictor, $\sigma(\cdot)$ is the Sigmoid function, $I_{\text{light}}$ and $I_{\text{depth}}$ provide visibility-related information, and $M_{\text{fo}}$ is the foreground object mask. Moreover, $G_{\text{p}}$ is jointly optimized with the second stage in an end-to-end manner, without additional explicit supervision, allowing it to adaptively identify challenging areas during the reconstruction process. However, the model might minimize $\mathcal{L}_{s2}$ by simply driving $S_{\text{prior}}$ toward a near-zero map. To avoid this gradient collapse, we apply mean normalization to enforce budget-preserving gradient reallocation:
\begin{equation}
\hat{S}_{\text{prior}}(x, y) = \frac{S_{\text{prior}}(x, y)}{\frac{1}{N} \sum_{i,j} S_{\text{prior}}(i, j)+\epsilon}.
\label{eq:norm}
\end{equation}
We retain the global basic supervision and add the spatially weighted correction term within $S_{\text{prior}}$:
\begin{equation}
\mathcal{L}_{swl} = \frac{1}{N} \sum_{x,y} \left( \hat{S}_{\text{prior}}(x, y) \cdot \ell_{\text{base}}(x, y) \right).
\label{eq:swl}
\end{equation}
SWL allocates more gradient budget to the error-prone shadow regions, thus improving boundary alignment and reducing shadow leakage. Thus, the entire loss for the second stage is:
\begin{equation}
  \mathcal{L}_{s2} = \mathcal{L}_{base} + \lambda_2 \mathcal{L}_{swl},
  \label{eq:s2}
\end{equation}

\subsubsection{Joint Loss}
Finally, we jointly fine-tune both stages using the following loss function. where $\lambda_1$, $\lambda_2$, and $\gamma$ are hyperparameters used to balance the different loss terms.
\begin{equation}
  \mathcal{L}_{joint} = \mathcal{L}_{s2} + \gamma \mathcal{L}_{s1},
  \label{eq:joint}
\end{equation} 
\section{Experiments}

To systematically assess the performance of VSDiffusion in shadow generation, we perform extensive quantitative comparisons, qualitative evaluations, ablation analyses, and limitations.

\begin{figure*}[!t]
  \centering
  \includegraphics[width=0.9\textwidth]{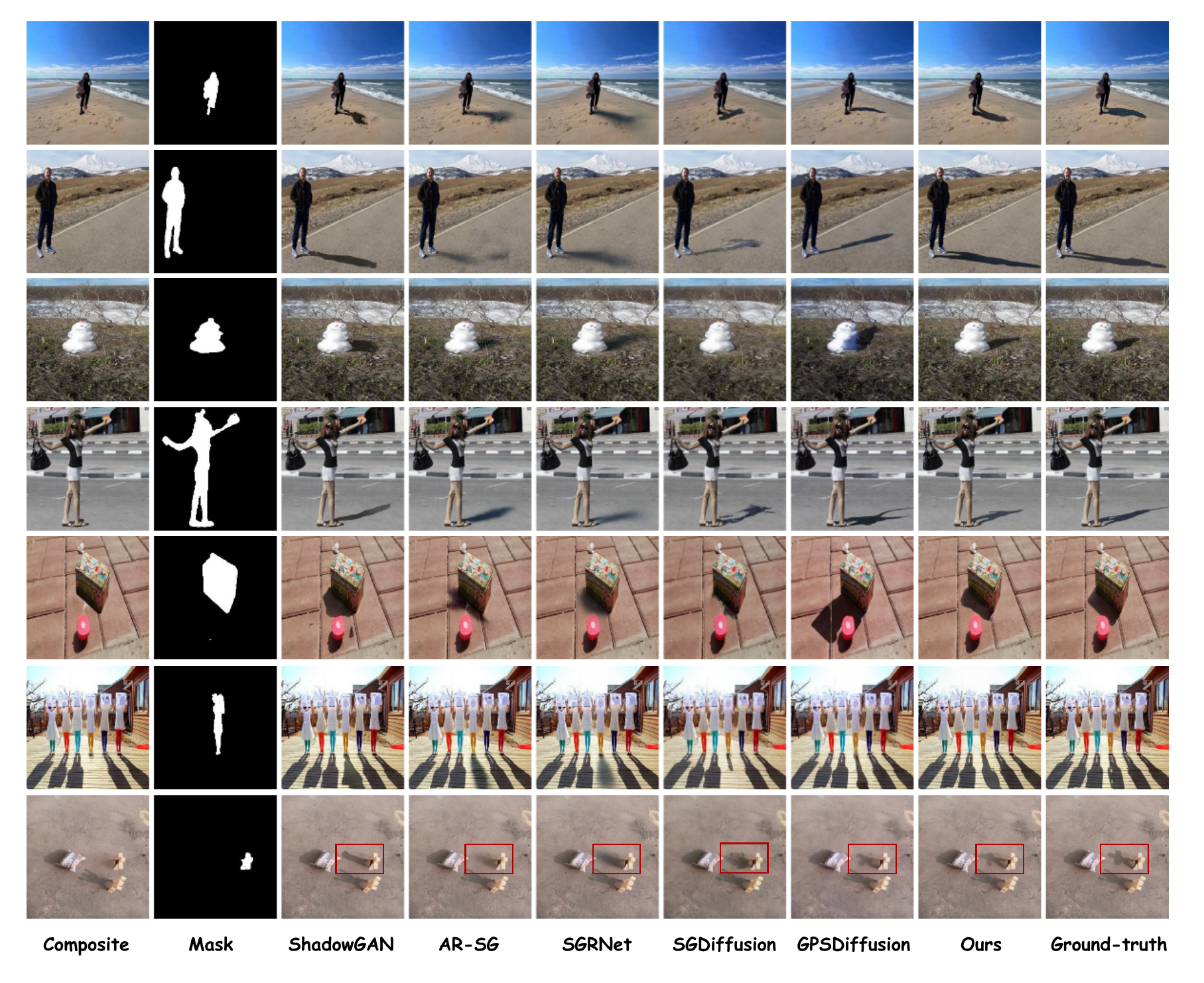}
  \caption{Qualitative comparison on the DESOBAv2. From left to right: the composite image (without foreground shadows), the foreground object mask, results of ShadowGAN \cite{zhang2019shadowgan}, AR-SG \cite{liu2020arshadowgan}, SGRNet \cite{hong2022shadow}, SGDiffusion \cite{liu2024shadow}, GPSDiffusion \cite{zhao2025shadow}, our VSDiffusion, and the ground truth. With visibility constraints, the proposed method generates shadows with more consistent projection direction, plausible contact, and sharper boundary alignment, especially under the BOS-free setting.
  }
  \label{fig:c1}
\end{figure*}

\subsection{Dataset and Implementation Details}

We conduct experiments on DESOBAv2, using the train/test split from \cite{liu2024shadow}, which contains 27,823 training images and 750 test images. The test set includes 500 BOS samples (with background object--shadow pairs) and 250 BOS-free samples. We evaluate both image quality and mask quality. For image quality, we report RMSE and SSIM over the full image, denoted as global RMSE (GR) and global SSIM (GS), respectively. We also compute local RMSE (LR) and local SSIM (LS) within the ground-truth foreground shadow region to better reflect shadow fidelity. For mask quality, we measure the balanced error rate (BER) between the ground-truth binary foreground shadow mask and the predicted mask binarized with a threshold of 0.5. We report global BER (GB) on the full image and local BER (LB) within the ground-truth foreground shadow region.

Our VSDiffusion is implemented in PyTorch and trained on two NVIDIA RTX 3090 GPUs with batch size 18 per GPU. We use Adam ($\beta_1{=}0.9$, $\beta_2{=}0.999$) for 700 epochs with an initial learning rate of $3\times10^{-5}$. We apply Kaiming initialization and EMA with decay 0.9999. All images are resized to $256\times256$ for both training and testing. Following \cite{guo2023shadowdiffusion}, we adopt an image-space conditional diffusion architecture to better preserve background textures for image-to-image shadow generation. Loss weights are set to $\lambda_1{=}0.2$, $\lambda_2{=}0.5$, and $\gamma{=}0.1$.

\subsection{Comparative experiments}

In this section, we mainly compared with representative non-rendering shadow generation methods, including ShadowGAN \cite{zhang2019shadowgan}, Mask-SG \cite{hu2019mask}, AR-SG \cite{liu2020arshadowgan}, SGRNet \cite{hong2022shadow}, DMASNet \cite{tao2024shadow}, SGDiffusion \cite{liu2024shadow}, and GPSDiffusion \cite{zhao2025shadow}. These baselines covered both one-stage and two-stage pipelines, as well as GAN-based and diffusion-based generators. 

\subsubsection{Quantitative comparison}

\cref{tab:quantitative_results} summarizes the quantitative results on DESOBAv2. Under both BOS and BOS-free settings, our method achieves the best or consistently competitive performance on the local metrics and BER, which are more sensitive to shadow geometry and boundary alignment. \textbf{BOS setting.} Compared with the current state-of-the-art method GPSDiffusion \cite{zhao2025shadow}, our approach reduces global BER (GB) and local BER (LB) by approximately 0.03 and 0.06, respectively, while achieving slight improvements in local RMSE (LR) and local SSIM (LS). \textbf{BOS-free setting.} When background object–shadow references are removed, most methods exhibit noticeable performance degradation, indicating increased ambiguity in the solution space. Despite this challenge, our method maintains stable performance, improving LS by about 0.02 and further reducing GB and LB by approximately 0.003 and 0.01, compared with the SOTA method. These results demonstrate that the proposed visibility-aware constraints and prior-guided supervision generalize effectively, enabling plausible shadow geometry and well-aligned boundaries even in the absence of explicit references.

%这一段对误码率的分析与之前的内容衔接的不太好。暂时删去。

%BER is sensitive to false positives (FPR) and false negatives (FNR). First, our coarse shadow region suppresses early diffusion into non-shadow areas, which reduces false positives. Next, the visibility constraint improves the plausibility of the projection direction and contact points, and reduces under-coverage. We also apply weighted supervision on boundary-critical regions to refine boundary alignment. These designs lead to consistent gains on BER. In contrast, the global metrics GR and GS can be dominated by large non-shadow regions. Thus, improving global appearance (\eg tone) may increase GR and GS, but it does not necessarily mean better shadow geometry. Our method focuses on geometric consistency and boundary alignment in the shadow region, so it achieves better performance on the local metrics and BER, which better match the core goal of this task.

\subsubsection{Qualitative comparison}

We evaluate VSDiffusion against other methods under both BOS-free (rows 1-4) and BOS (rows 5-7) settings in \cref{fig:c1}. While existing diffusion-based models achieve basic darkening, they often struggle with shadow direction, shape fidelity, and boundary alignment. For instance, in row 2, SGDiffusion \cite{liu2024shadow} and GPSDiffusion \cite{zhao2025shadow} show direction errors, where the generated shadows are inconsistent with the scene lighting direction. In row 5, AR-SG \cite{liu2020arshadowgan} produces distorted shadow shapes, and the contours do not match the object geometry. In addition, SGRNet \cite{hong2022shadow} often produces blurred edges and halo-like transitions across multiple examples, which reduces local boundary accuracy. In contrast, VSDiffusion generates shadows with more consistent geometry and sharper boundaries across diverse scenes, and the advantage is more evident under BOS-free. This work introduces visibility priors to constrain shadow direction and contact relationships, which narrows the feasible solution space even without reference cues. It further applies the $S_{\text{prior}}$-Weighted Loss (SWL) to emphasize boundary-critical regions and uses High-Frequency Guided Enhancement (HFGE) to enhance high-frequency details, leading to more accurate contours and boundaries.

\subsection{Ablation Study}

\begin{table}[!t]
  \caption{Ablation study on DESOBAv2. VCB and SWL are two visibility-inspired ways to inject priors into the diffusion model. SWL is short for the $S_{\text{prior}}$-Weighted Loss. HFGE denotes the High-Frequency Guided Enhancement module. $\uparrow$/$\downarrow$ indicate higher/lower is better, and best results are in bold.
  }
  \label{tab:ab1}
  \centering
  \setlength{\tabcolsep}{3pt} % 和参考表格完全一致的列间距
  \small % 单栏表格标准字号，和参考表格字体大小匹配
  \begin{tabular}{@{}ccc|cccccc@{}}
    \toprule
    VCB & HFGE & SWL & GR$\downarrow$ & LR$\downarrow$ & GS$\uparrow$ & LS$\uparrow$ & GB$\downarrow$ & LB$\downarrow$ \\
    \midrule
        &      &      & 6.655 & 50.598 & 0.962 & 0.364 & 0.221 & 0.438 \\
    \checkmark &      &      & 6.592 & 48.265 & 0.963 & 0.366 & 0.203 & 0.403 \\
    \checkmark & \checkmark &      & 6.571 & 48.147 & \textbf{0.964} & 0.374 & 0.193 & 0.381 \\
    \checkmark & \checkmark & \checkmark & \textbf{6.422} & \textbf{46.542} & \textbf{0.964} & \textbf{0.377} & \textbf{0.182} & \textbf{0.360} \\
    \bottomrule
  \end{tabular}
\end{table}

\begin{figure}[!t]
  \centering
  \includegraphics[width=\linewidth]{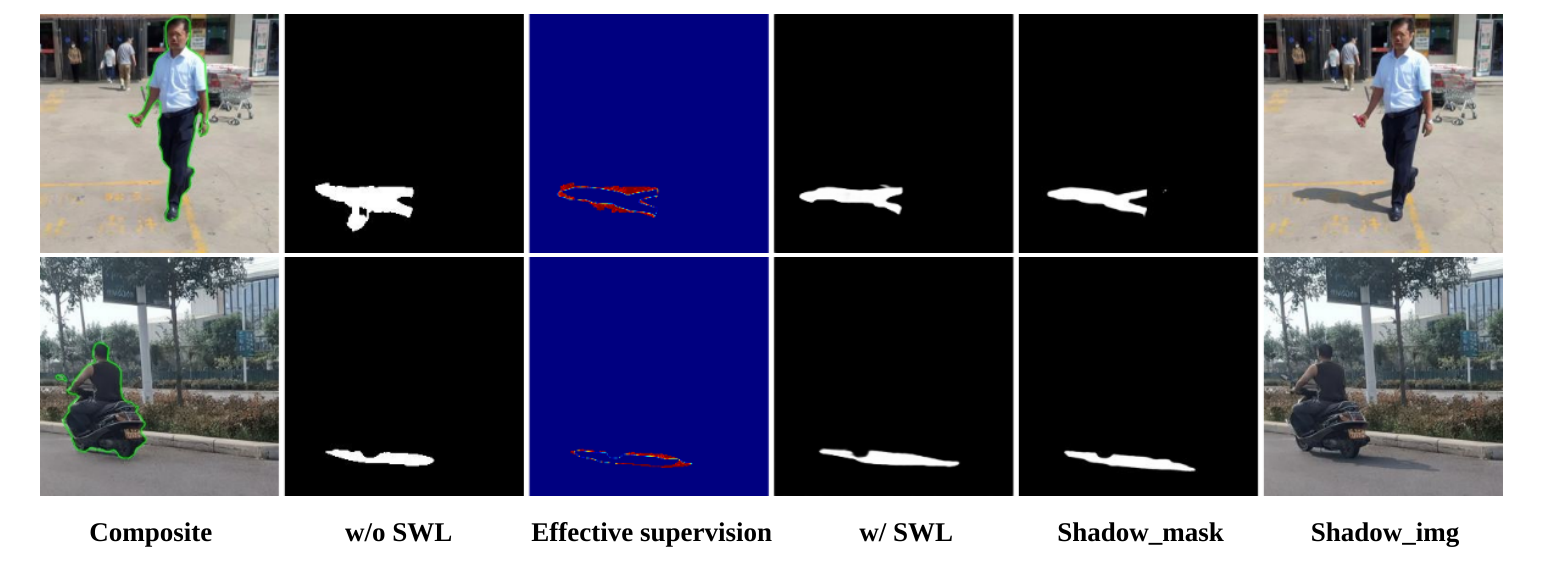}
  \caption{Qualitative ablation of VSDiffusion. Visual comparisons demonstrate that our SWL effectively concentrates supervision on critical shadow boundaries and penumbras. In Effective supervision, the red regions \ie, high weight $\times$ high error.
  }
  \label{fig:ab1}
\end{figure}

As shown in \cref{tab:ab1}, we evaluate the effectiveness of each component on the DESOBAv2 dataset. Overall, the results indicate that both visibility-aware priors and high-frequency refinement play essential roles in improving shadow geometry consistency and boundary alignment. 

\noindent 1) Visibility Priors (VCB \& SWL). The Visibility Control Branch (VCB) and $S_{\text{prior}}$-Weighted Loss (SWL) incorporate visibility guidance through complementary structural conditioning and adaptive supervision. Specifically, VCB imposes architectural constraints to regularize shadow placement, thereby improving geometry-related metrics such as BER and LR. 
SWL reshapes the training signal in a spatially adaptive manner. As visualized in \cref{fig:ab1} (col.~3), we show the effective supervision map $\hat{S}_{\text{prior}} \cdot \ell_{\text{base}}$. Here, red regions correspond to \emph{high weight $\times$ high error},  
indicating that SWL emphasizes supervision on difficult regions, typically around shadow boundaries and penumbras. Consequently, with SWL (col.~4), the refined shadow mask exhibits better boundary alignment with the ground-truth mask (col.~5) than the setting without SWL (col.~2). Quantitatively, SWL further reduces LR by 1.605 and LB by 0.021 compared with the third row in \cref{tab:ab1}.

\noindent 2) High-Frequency Refinement (HFGE): The HFGE module improves local structural fidelity by leveraging high-frequency encoder representations. By enhancing fine-grained geometric cues, HFGE suppresses texture-level distortion and improves Local SSIM (LS) by 0.008 over the second row.

\begin{table}[!t]
\centering
\caption{Ablation on light/depth priors in VCB on DESOBAv2. 
“w/o light” and “w/o depth” remove the corresponding condition while keeping all other components unchanged. 
The full model achieves the best overall performance. 
$\downarrow$/$\uparrow$ indicate lower/higher is better, and best results are in bold.}
\label{tab:ab2}
\setlength{\tabcolsep}{5.5pt}
\small
\begin{tabular}{l|cccccc}
\toprule
Method & GR$\downarrow$ & LR$\downarrow$ & GS$\uparrow$ & LS$\uparrow$ & GB$\downarrow$ & LB$\downarrow$ \\
\midrule
Baseline   & 6.655 & 50.598 & 0.962 & 0.364 & 0.221 & 0.438 \\
w/o light  & 6.545 & 48.149 & 0.963 & 0.373 & 0.197 & 0.390 \\
w/o depth  & 6.456 & 47.990 & 0.963 & 0.371 & 0.187 & 0.369 \\
Ours       & \textbf{6.422} & \textbf{46.542} & \textbf{0.964} & \textbf{0.377} & \textbf{0.182} & \textbf{0.360} \\
\bottomrule
\end{tabular}
\end{table}
Our earlier visibility analysis shows that shadow geometry is mainly determined by the spatial relationship among the light source, the caster, and the shadow receiver. Therefore, VCB uses two conditions, \ie, Light and Depth. To verify whether these two conditions are redundant, we remove each condition separately under the same training setup. As shown in \cref{tab:ab2}, removing either condition degrades performance, while using both gives the best results. This shows that Light and Depth are complementary rather than redundant: Light constrains illumination geometry, while Depth provides scene-structure cues. Together, they more effectively shrink the feasible set of shadow solutions.

\section{Limitations and Future Work}
\label{sec:Limit}
\begin{figure}[!h]
  \centering
  \includegraphics[width=0.98\linewidth]{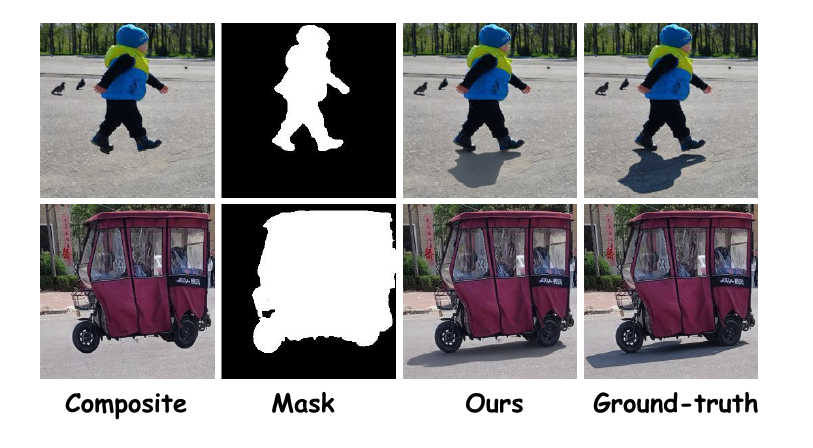}
  \caption{Limitations in BOS-free scenarios: our shadows match geometry but may slightly deviate in intensity from the ground truth.
  }
  \label{fig:l1}
\end{figure}

Despite our VSDiffusion generating geometrically consistent shadows in most cases, we observe slight shadow-intensity inconsistency in a small number of BOS-free scenarios, without a background reference. As shown in \cref{fig:l1}, under the BOS-free setting, our result matches the ground truth well in shadow direction and shape, while the shadow intensity can be slightly underestimated. We attribute this issue mainly to insufficient cues under the BOS-free setting: the input $M_{fo}$ provides only a contour-level hint of the foreground and contact area, while lacking material-related signals, such as reflectance, translucency and local occlusion cues that are important to stably calibrate shadow opacity and intensity.

In future work, we plan to extend our framework to photorealistic subject-driven image editing. To further improve realism in BOS-free scenarios, we will introduce a background-reference-free adaptive calibration mechanism to better regulate shadow intensity while preserving geometric consistency. 
\section{Conclusion}

In this paper, we present VSDiffusion for physically consistent shadow generation. We formulate shadow generation as an inherently ill-posed problem and reduce its ambiguity by explicitly modeling the visibility formation process to progressively constrain the solution space. Our two-stage framework combines coarse shadow localization with diffusion refinement guided by visibility priors derived from light, caster, and shadow receiver modeling. Structurally controllable conditioning and adaptive spatial supervision serve as complementary mechanisms that enable effective prior integration, while a high-frequency enhancement module improves boundary sharpness and texture fidelity. Extensive experiments demonstrate stable improvements in geometric alignment and visual realism, particularly under reference-free settings. These results highlight the effectiveness of visibility-derived constraints in reducing ambiguity in shadow generation. 

{
    \small
    \bibliographystyle{ieeenat_fullname}
    \bibliography{main}
}

\end{document}